\documentclass[runningheads]{llncs}
\usepackage[T1]{fontenc}
\usepackage{graphicx}
\usepackage{amssymb}
\usepackage{amsmath}
\usepackage{booktabs}
\usepackage{subcaption}
\usepackage{orcidlink}

\usepackage{fancyhdr}
\fancypagestyle{firstpage}{
    \fancyhf{} % Clear default headers and footers
    \fancyhead[C]{\scriptsize Cite as: ``López, F. M., \& Cruz, F. (2026). Can Hindsight Critics Guide Delayed Actors?. In Australasian Joint Conference on Artificial Intelligence (pp. 1-10).''} % C=Center, L=Left, R=Right
    \fancyfoot[C]{\thepage} % Retain the page number at the bottom
}

\begin{document}

\title{
Can Hindsight Critics Guide Delayed Actors?
}

\titlerunning{
Can Hindsight Critics Guide Delayed Actors?
}

\author{
Francisco M. López\inst{1,2,\orcidlink{0000-0002-9653-1740}} \and
Francisco Cruz\inst{1,3,\orcidlink{0000-0002-1131-3382}}
}
\authorrunning{F. M. López and F. Cruz}
\institute{
School of Computer Science and Engineering, UNSW Sydney, Australia \and
Frankfurt Institute for Advanced Studies, Germany \and
Escuela de Ingeniería, Universidad Central de Chile, Chile\\
\email{
fm.lopez@unsw.edu.au,
f.cruz@unsw.edu.au}
}

\maketitle              % typeset the header of the contribution
%

% ARXIV HEADER
\thispagestyle{firstpage}

\begin{abstract}

Sensory delays create temporal misalignments between the state of the environment and the observations available to an agent, turning Markovian control problems into partially observable ones. In this work, we inquire whether such misalignments can be mitigated during training by using hindsight to recover aligned state-action transitions. We introduce \emph{Hindsight Critics}, an asymmetric extension of the Soft Actor-Critic for reinforcement learning under partial observability. The critics are trained on hindsight-aligned transitions whereas the actor learns from the resulting privileged value estimates despite only having access to delayed observations. Comparisons against delay-aware and unaware baselines on the DelayBench suite show that Hindsight Critics can partially compensate for observation delays without requiring modifications to the agent's observations during deployment.

\keywords{Reinforcement Learning \and Partial Observability \and Sensory Delays \and Soft Actor-Critic \and Asymmetric Actor-Critic \and Hindsight} 

\end{abstract}
%

% 
% --- Main body ---
% 

\section{Introduction}

Observation delays are ubiquitous for biological and robotic agents \cite{whelan2008effective,codol2023sensorimotor}. The latency between sensing and decision-making can be large enough that the world's state has shifted by the time an action is selected \cite{neto2026reinforcement}. From the perspective of reinforcement learning (RL), delays change the statistical structure of the control problem faced by the algorithm. In a delay-free Markov Decision Process (MDP) \cite{sutton1998reinforcement}, the environment's state \(s_t\) connects the policy input, $a_t \sim \pi(\cdot|s_t)$, the state transitions, $s_{t+1}\sim p(\cdot|s_t,a_t)$, and the collected rewards, $r_t=r(s_t,a_t)$.  With an observation delay \(\tau\), this coupling breaks down. The agent observes a past state $s_{t-\tau}$ and must select an action from delayed information, $a_t~\sim~\pi(\cdot|s_{t-\tau})$, while the reward and transition are still generated by the current state-action pair \((s_t,a_t)\). The resulting mismatch between the agent's decision and the environment can be modelled as a Partially Observable MDP (POMDP) \cite{sutton1998reinforcement}.  

Standard RL algorithms fail to compensate for observation delays \cite{delaybench}. It is common practice to attempt to solve POMDPs by solving closely-related (approximately) Markovian problems \cite{neto2026reinforcement}. For example, delay-aware algorithms typically rely on history-based representations \cite{kaelbling1998planning}, recurrent policies \cite{nath2021revisiting,hausknecht2015deep,pleines2022generalization}, or forward models \cite{karamzade2024reinforcement,karamzade2025model,wu2025directly}. Delay-aware approaches can successfully mitigate some of the consequences of delays by modifying the information available to the deployed policy. Yet such modifications may be difficult or impractical for real-world deployments \cite{neto2026reinforcement,dulacarnold2020realworldrlempirical}. This leads us to the central question of this study: can we improve learning under observation delays without modifying the input to the agent's policy?

In this paper, we introduce \emph{Hindsight Critics}, an asymmetric extension of the Soft Actor-Critic (SAC) \cite{haarnoja2018soft} for RL in POMDPs. The actor remains constrained to the partial observations available at deployment. However, rollout transitions are stored in a replay buffer where the true environment's state can be estimated from hindsight. The critics are trained with these hindsight transitions, while the actor is updated through the resulting value estimates despite only having access to the partial observations. Thus, hindsight information guides policy learning during training without being provided to the actor at deployment. We evaluate our method on the DelayBench suite of continuous control tasks against delay-unaware and delay-aware baselines. The results show that Hindsight Critics can improve learning under observation delays while preserving the same delayed observation interface at deployment.

\section{Related Work}

\subsection{Asymmetric actor-critics under partial observability}

The method presented here belongs to the family of asymmetric actor-critic methods \cite{pinto2017asymmetric,baisero2021unbiased,lambrechts2025theoretical}, in which actors and critics are conditioned on different information during training. These methods are common in robotics, where privileged full state information from simulations is used to train critics while actors only receive partial observations from cameras or sensor readings \cite{pinto2017asymmetric}. More generally, asymmetric actor-critics exploit the fact that critics need not be deployed: they are only used during training and thus can use privileged information that is unavailable to the actors, provided that the policy itself does not use that privileged information \cite{baisero2021unbiased,lambrechts2025theoretical,pinto2017asymmetric}. Our method follows this asymmetric principle but applies it to the temporal structure of observation delays.

\subsection{Compensating for observation delays}

Observation delays constitute a structured form of partial observability. The agent receives a full description of the environment, but one that corresponds to a past state rather than to the current one where actions are executed. Delays have long been recognised as a practical obstacle for control systems and embodied agents, where sensing, communication, computation, and actuation can all introduce latency issues \cite{katsikopoulos2003markov,neto2026reinforcement}. In RL, delays break the usual temporal correspondence between observations, actions, and rewards. Prior work has addressed this problem by augmenting the agent's state with recent observations and actions, thereby transforming delayed problems into higher-dimensional Markovian problems when the delay is known or bounded \cite{nath2021revisiting}. Other approaches use forward or world models to predict the current state from delayed observations, reducing delayed POMDPs to approximately delay-free control problems \cite{wu2025directly,karamzade2024reinforcement,karamzade2025model}. Recurrent policies follow a similar motivation by providing the policy with enough temporal context to infer the current control state \cite{hausknecht2015deep,pleines2022generalization}. In practice, however, these solutions increase the complexity of the policy, making memory-dependent control inefficient for fast deployments. Our work inquires whether the learning signal can be improved using hindsight while leaving the actor's delayed observation unchanged.

\subsection{Learning from hindsight in replay buffers}

Hindsight Experience Replay (HER) is a popular method that relabels goals in sparse-reward goal-conditioned tasks \cite{andrychowicz2017hindsight}. We use hindsight not to relabel goals or rewards, but to restore temporal alignment at the end of each episode. Closest to our work, Delay-Correcting Actor-Critic (DCAC) addresses random action and observation delays by deriving delay-corrected multi-step targets through partial trajectory resampling \cite{bouteiller2020reinforcement}. DCAC showed that replay buffers contain useful chronological structure that can be exploited through hindsight. Our method exploits this hindsight perspective differently: we align replayed transitions and use them exclusively to train the critics, such that the actor remains constrained to the delayed observation available at deployment.

\section{Hindsight Critics}

\subsection{Problem setup}

Let us consider the problem of RL in a Markovian environment under a fixed observation delay $\tau$. At time $t$, the environment is in state $s_t$ but the agent observes only a delayed state: 
\begin{equation}
    x_t = s_{t-\tau},
\end{equation}
and selects actions according to its (delayed) policy
\begin{equation}
    a_t \sim \pi(\cdot \mid x_t).
\end{equation}

The action is applied to the current environment state, which results in a Markovian transition and reward,
\begin{equation}
    s_{t+1} \sim p(\cdot \mid s_t, a_t), \quad r_t = r(s_t, a_t).
\end{equation}

The objective is to learn a deployable policy \(\pi(\cdot \mid x_t)\) that maximises the expected discounted return, $R_t = \sum_k \gamma^k r_{t+k}$. However, because the policy is conditioned on the delayed observation $x_t=s_{t-\tau}$ rather than the current state $s_t$, the agent's decision is non-Markovian with respect to the reward and transition. In sum, the agent faces a POMDP with delayed observations. Delay-unaware methods attempt to solve the POMDP using the default stored transitions, $(x_t, a_t, r_t, x_{t+1}, \mathrm{done}_t)$, leading to fragile performances that struggle even with moderate delays \cite{delaybench}.

\subsection{Transitions aligned in hindsight}

Although the agent must act from delayed observations, the chronological structure of a completed trajectory allows the transition associated with each action and reward to be recovered in hindsight. At the end of an episode, the replay buffer can align the observation $x_t$ with an estimate of the state of the environment $\tilde{s}_t$ to generate hindsight-aligned transitions:
\begin{equation}
(x_t, \tilde{s}_t, a_t, r_t, x_{t+1}, \tilde{s}_{t+1}).
\end{equation}

Under the assumption of fixed delay $\tau$, the state estimates are obtained directly from future observations
\begin{equation} \label{eq:state_estimate}
\tilde{s}_t = x_{t+\tau},
\quad
\tilde{s}_{t+1} = x_{t+\tau+1}.
\end{equation}

That is, delays can be compensated with hindsight by using a later observation from the same trajectories. Boundary cases at episode termination are handled using the terminal state returned by the environment. The hindsight-aligned transition format preserves the deployable observations $x_t$ while also providing an estimate of the current state $\tilde{s}_t$ associated with each state-action transition and reward.

\subsection{Soft Actor-Critic with Hindsight Critics}

Hindsight Critics exploit the hindsight-aligned transitions within an asymmetric variant of the Soft Actor-Critic (SAC) \cite{haarnoja2018soft}. The agent consists of one actor $\pi_\theta$ and two critics $Q_{\phi}$ and $Q_{\psi}$. The asymmetry stems from the different information the actor and the critics are trained on. 

The actor remains unchanged with respect to the POMDP, and thus remains deployable with delayed observations,
\begin{equation} \label{eq:critics}
\pi_\theta(a_t \mid x_t),
\end{equation}
whereas the twin critics are trained on the hindsight-aligned transitions using the current state estimates,
\begin{equation}
Q_{\phi}(x_t, \tilde{s}_t,a_t),
\qquad
Q_{\psi}(x_t, \tilde{s}_t,a_t).
\end{equation}

\subsubsection{Critics training:}

The critics update follow the standard entropy-regularised SAC target, except that the next action is sampled from the delayed actor input $x_{t+1}$,
\begin{equation}
\hat{a}_{t+1} \sim \pi_\theta(\cdot \mid x_{t+1}),
\end{equation}
while the target critics $Q_{\bar{\mu}}$ for $\mu\in \{\phi,\psi\}$,  are evaluated at the hindsight-aligned next state estimates $\tilde{s}_{t+1}$:
\begin{equation}
y_t =
r_t + \gamma \, (1-\mathrm{done}_t)
\left[
\min_{\mu} Q_{\bar{\mu}}(x_{t+1}, \tilde{s}_{t+1},\hat{a}_{t+1})
-
\alpha \log \pi_\theta(\hat{a}_{t+1}\mid x_{t+1})
\right].
\end{equation}

The critic loss becomes:
\begin{equation}
\mathcal{L}_Q
=
\sum_{\mu\in\{\phi,\psi\}}
\mathbb{E}_{\mathcal{D}}
\left[
\left(
Q_{\mu}(x_t, \tilde{s}_t,a_t) - y_t
\right)^2
\right].
\end{equation}

\subsubsection{Actor training:}

On the other hand, the actor is updated from delayed observations, but its sampled actions
\begin{equation}
a_t^\pi \sim \pi_\theta(\cdot \mid x_t)
\end{equation}
 are evaluated by the hindsight critics:
\begin{equation}
\mathcal{L}_\pi
=
\mathbb{E}_{\mathcal{D}}
\left[
\alpha \log \pi_\theta(a_t^\pi \mid x_t)
-
\min_\mu Q_{\mu}(x_t, \tilde{s}_t,a_t^\pi)
\right].
\end{equation}

In sum, the actor receives policy-gradient information from critics trained on an estimate of the current state, but the actor itself is never conditioned on that state. During deployment rollouts and evaluations, the critics are discarded and the actor's policy chooses actions exclusively from the delayed observations.

We hypothesise that the asymmetry introduced by Hindsight Critics is sufficient to mitigate the consequences of the partial observability of delayed observations without directly removing the partial observability. Instead, our method addresses the consequences of delays from estimated state-action transitions. In this sense, hindsight information guides actor learning during training without changing the information available to the actor at deployment.

\section{Experiments}

We hypothesise that Hindsight Critics can improve RL under observation delays while keeping the delayed policy unchanged. Our experiments are designed to assess how the resulting performance compares to standard delay-aware and delay-unaware RL algorithms.

\subsection{Benchmark and environments}

We use the DelayBench suite \cite{delaybench}, a benchmark that augments standard Gymnasium continuous control environments with controlled observation delays. At time $t$, the agent receives $x_t=s_{t-\tau}$, while the environment transition and reward are generated by the current state-action pair $(s_t,a_t)$. We evaluate policies over the default range observation delays
\begin{equation}
\tau \in \{0,1,2,4,8,16\} \; \mathrm{steps} ,
\end{equation}
across 8 different environments ranging from classic control to MuJoCo locomotion. We use the default observation and action spaces, reward structures, episode lengths, and training budgets for each environment as set by default in the DelayBench suite. 

\subsection{Algorithms}

Our main contribution is Hindsight Critics, an asymmetric extension of SAC. The actor receives only the delayed observation whereas the critics are trained with the addition of the hindsight-aligned states:
\begin{equation}
a_t \sim \pi_\theta(\cdot \mid x_t),
\quad 
Q_{\mu}(x_t,\tilde{s}_t,a_t).
\end{equation}

\subsubsection{Delay-unaware algorithms}

We compare Hindsight Critics with other standard RL algorithms: PPO, A2C, DDPG, TD3, and SAC. These algorithms treat the delayed environment as an ordinary MDP and do not explicitly compensate for the temporal misalignment. Thus, unlike Hindsight Critics, these algorithms rely on the delayed observations for both actors and critics:
\begin{equation}
a_t \sim \pi(\cdot\mid x_t),
\quad
Q(x_t,a_t).
\end{equation}
In particular, the comparison with SAC isolates the effect of the additional state estimate input to the critics, since the algorithms are otherwise identical. 

\subsubsection{Delay-aware algorithms}

We also compare Hindsight Critics with the delay-aware variants of PPO and SAC provided in the DelayBench suite. Concretely, we consider two delay mitigation strategies:
\begin{itemize}
    \item \textbf{History stacking}, where the input is augmented with a concatenation of the actions selected by the agent during the delay window:
    \begin{equation} \label{eq:history}
        x_t' = x_{t} \oplus a_{t-\tau} \oplus a_{t-\tau+1} \oplus ... \oplus a_{t-1}
    \end{equation}
    \item \textbf{Forward model}, where the input is augmented with a prediction of the true state from the delayed state and the agent's actions:
    \begin{equation} \label{eq:forward}
        x_t' = x_{t} \oplus f(x_{t}, a_{t-\tau}, a_{t-\tau+1}, ..., a_{t-1})
    \end{equation}
\end{itemize}

\subsection{Evaluation metrics}

Our evaluation protocol follows the guidelines of the DelayBench suite \cite{delaybench}. For each algorithm, environment, and delay, a policy is trained with 20 random seeds and evaluated deterministically over 20 evaluation episodes. The results reported in Section \ref{sec:results} are computed from the averages over those $20~\times~20~=~400$ evaluation episodes per training condition. 

To compare performances, we first compute the average evaluation returns per condition:
\begin{equation}
G_{\tau}
=
\frac{1}{NM}
\sum_{i=1}^{N}
\sum_{j=1}^{M}
G_{\tau}^{(i,j)},
\end{equation}
where $N=20$ is the number of training seeds and $M=20$ is the number of evaluation episodes per seed. Because returns are observed only at a discrete set of delays, we fit a decreasing sigmoid $\hat{G}(\tau)$ to the return-delay curve. We then summarise robustness across delays by means of the area-under-the-curve (AUC) score:
\begin{equation}
    \mathrm{AUC} =  \frac{1}{\tau_m} \int_{0}^{\tau_m} \mathrm{ReLU}\left( \frac{\hat{G}_\tau - G_{\mathrm{random}}}{G_{max} - G_{\mathrm{random}}}\right) \,d\tau \,,
\end{equation}
where the integral is performed between the delays of 0 and $\tau_m=16$ timesteps, $G_{max}$ is the highest return obtained from a single evaluation episode across all algorithms and delays that serves as the maximum baseline, and $G_\mathrm{random}$ is the return of a random policy that serves as the minimum baseline (see \cite{delaybench}). The return-delay curves measure how the average returns change as the observation delays increase. The AUC has values from 0 to 1, with higher scores indicative of both strong zero-delay performance and slower degradation under increasing delay.

\begin{figure}[!t]
     \centering
     \begin{subfigure}[b]{0.32\textwidth}
         \centering
         \includegraphics[width=\linewidth]{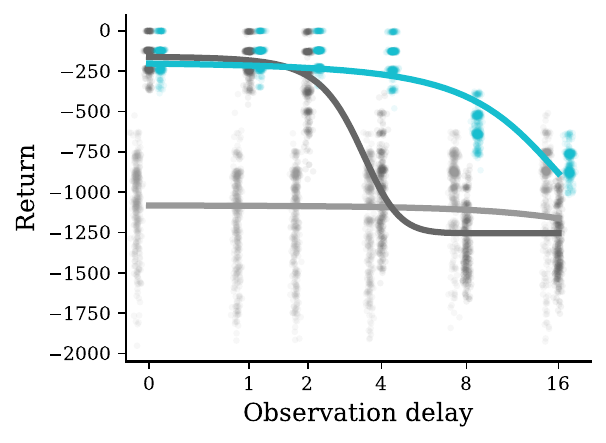}
         \caption{\texttt{Pendulum-v1}}
     \end{subfigure}
     \hfill
     \begin{subfigure}[b]{0.32\textwidth}
         \centering
         \includegraphics[width=\linewidth]{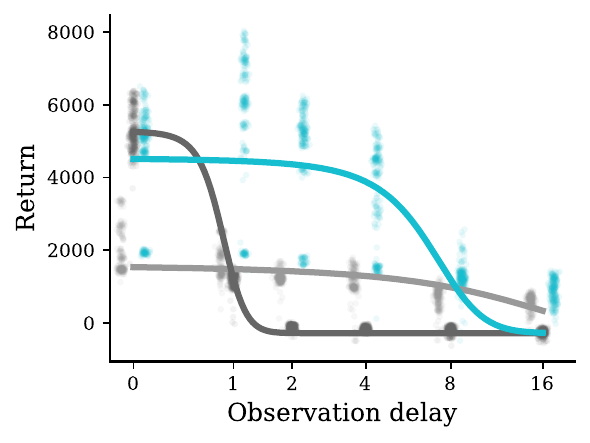}
         \caption{\texttt{HalfCheetah-v5}}
     \end{subfigure}
     \hfill
     \begin{subfigure}[b]{0.32\textwidth}
         \centering
         \includegraphics[width=\linewidth]{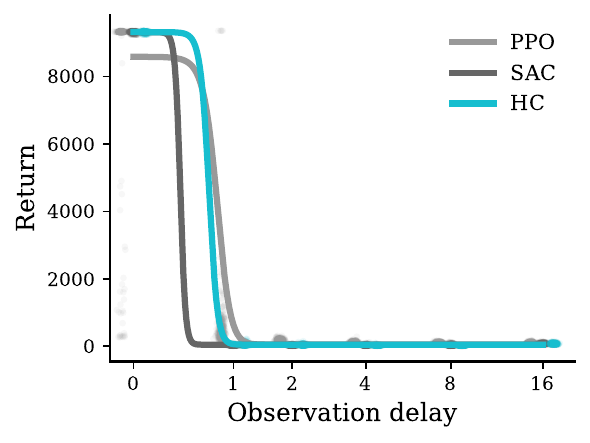}
         \caption{\texttt{Inv.D.Pend.-v5}}
     \end{subfigure}
     \caption{Return-delay curves for PPO, SAC and Hindsight Critics (HC) over three illustrative environments. Scattered dots indicate the return over a single evaluation episode. Sigmoid curves are fitted to those returns and used to compute the AUC scores. SAC achieves high zero-delay performances but is particularly sensitive to delays. Hindsight Critics can partially mitigate this degradation.}
     \label{fig:hc_vs_sac}
\end{figure}

\section{Results} \label{sec:results}

\subsection{Hindsight Critics can match SAC in zero-delays}

First, we compare Hindsight Critics against common delay-unaware RL algorithms in a zero-delay configuration. The results presented in Table~\ref{tab:zerodelay} show that Hindsight Critics performs comparably to SAC and better than the other algorithms, with only DDPG and PPO achieving the highest returns in 2 environments each. The similarity with SAC is unsurprising, since both algorithms share the same architecture and loss functions. The only difference is in the input to the critics. Under zero-delays, the observation $x_t$ and the state estimate $\tilde{s}_t$ are the same (see Eq.~\ref{eq:state_estimate}), and thus the critics of Eq.~\ref{eq:critics} receive the observation $x_t$ twice as an input, whereas the SAC critics only receive it once. Nevertheless, the finding that Hindsight Critics can match and occasionally outperform SAC under zero-delays serves as a validation of our method beyond its original purpose of compensating for sensory delays.

\begin{table}[!t]
\caption{Zero-delay average returns of Hindsight Critics (HC) compared with other common RL algorithms across multiple environments. The highest score for each environment is highlighted in \textbf{bold}.} 
\label{tab:zerodelay}
\centering
\begin{tabular*}{\textwidth}{l@{\extracolsep{\fill}}cccccc}
\\
\toprule
Environment     & PPO & A2C & DDPG & TD3 & SAC & \textbf{HC}\\
\midrule
\texttt{Pendulum-v1} & -1068.7 & -1245.3 & \textbf{-139.3} & -140.9 & -156.7 & -154.8 \\
\texttt{InvertedPendulum-v5}  & \textbf{1000.0} & 856.8 & 761.8 & 954.4 & \textbf{1000.0} & \textbf{1000.0}\\
\texttt{InvertedDoublePendulum-v5}  & 8589.9 & 501.8 & 7496.3 & 8616.9 & \textbf{9319.8} & \textbf{9319.8} \\
\texttt{Reacher-v5}  & -4.7 & -26.8 & \textbf{-4.5} & -4.7 & -5.6 & -5.7 \\
\texttt{HalfCheetah-v5} & 1942.3 & 813.8 & 4235.7 & --  & \textbf{5282.5} & 4571.6\\
\texttt{Hopper-v5}  & 1964.0 & 148.0 & 873.6 & 89.6 & \textbf{3247.0} & 2630.4 \\
\texttt{Walker2d-v5} & 1528.6 & 114.3 & 969.6 & 1310.3 & 3581.5 & \textbf{3940.2} \\
\texttt{Swimmer-v5}  & \textbf{66.9} & 25.2 & 37.7 & 40.2 & 46.3 & 50.9\\
\bottomrule
\end{tabular*}
\end{table}

\subsection{Hindsight Critics can partially mitigate delay degradation}

Table~\ref{tab:unaware} shows the AUC scores obtained from training each algorithm across a wide range of observation delays. Despite all actors being delayed, i.e. not solving the problem of partial observability, we find strong differences in the way delays influence the performance of each algorithm. Out of the 8 environments used in this study, Hindsight Critics achieves the highest score in 4 of them, followed by PPO with 3 and DDPG with 1. TD3 and SAC achieve mid-range scores across all environments whereas A2C is consistently poor.

The robustness of PPO to observation delays had already been identified in \cite{delaybench} and is aligned with a large body of research showing that PPO is consistently the most robust algorithm under changing environment conditions and other perturbations \cite{larsen2021comparing}. This observation is typically attributed to the clipping of the actor objective used by PPO's on-policy learning. By comparison, off-policy methods such as SAC or TD3 achieve high returns without perturbations but are strongly affected by delays, resulting in lower AUC scores. This result explains our decision to use SAC as the backbone architecture for Hindsight Critics: strong zero-delay performance which can most benefit from increased robustness to delays.

Table~\ref{tab:unaware} shows that Hindsight Critics can partially solve SAC's degradation exclusively by modifying the critics but without the need to modify the input to the deployed policy. Fig.~\ref{fig:hc_vs_sac} shows the return-delay curves for SAC and Hindsight Critics over three illustrative environments. We find that both methods achieve similar performances without delays. However, SAC has a strong performance drop even with relatively small delays of 4 timesteps, whereas Hindsight Critics is able to partially mitigate this degradation. On the other hand, with large delays of 16 timesteps, Hindsight Critics also  struggles to learn satisfactory behaviours. 

\begin{table}[!t]
\caption{AUC scores of Hindsight Critics (HC) compared with other common RL algorithms across multiple environments. The highest score for each environment is highlighted in \textbf{bold}.} 
\label{tab:unaware}
\centering
\begin{tabular*}{\textwidth}{l@{\extracolsep{\fill}}cccccc}
\\
\toprule
Environment     & PPO & A2C & DDPG & TD3 & SAC & \textbf{HC}\\
\midrule
\texttt{Pendulum-v1} & .112 & .000 & .192 & .205 & .182 & \textbf{.634} \\
\texttt{InvertedPendulum-v5}  & \textbf{.215} & .028 & .013 & .019 & .023 & .024\\
\texttt{InvertedDoublePendulum-v5}  & .035 & .002 & .038 & .034 & .037 & \textbf{.043} \\
\texttt{Reacher-v5}  & .701 & .374  & \textbf{.725} & .597 & .509 & .490 \\
\texttt{HalfCheetah-v5}  & .144 & .010 & .041 & -- & .034 & \textbf{.237}\\
\texttt{Hopper-v5} & \textbf{.095} & .007 & .027 & .010 & .034 & .030\\
\texttt{Walker2d-v5} & \textbf{.102} & .015 & .015 & .050  & .013 & .033\\
\texttt{Swimmer-v5}  & .313 & .173 & .169 & .171 & .198 & \textbf{.322}\\
\bottomrule
\end{tabular*}
\end{table}

\subsection{Hindsight Critics can complement other delay-aware methods}

Having found that hindsight can partially alleviate the performance degradation of delay-unaware algorithms, we inquire whether our algorithm can be combined with other delay-aware methods. Concretely, we compare PPO, SAC, and Hindsight Critics trained with history stacking and a forward model. The results are shown in Table~\ref{tab:aware}. Like before, we find that Hindsight Critics outperforms the other algorithms on a majority of the environments (5 out of 8 with history stacking, 4 out of 8 with the forward model). That is, despite the fact that the partial observability is virtually solved in these experiments, hindsight remains useful to mitigate the consequences of the delayed observations.

A comparison of Tables~\ref{tab:unaware}~and~\ref{tab:aware} indicates that all methods increase their AUC scores when the inputs to the actors are augmented with history stacking or forward model predictions. These increments require increasing the policy's input (see Eqs.~\ref{eq:history}~and~\ref{eq:forward}) as well as additional computations during deployment: history stacking necessitates a memory of recent actions whereas the forward model is a separate network that needs to be deployed together with the agent. Interestingly, the boosts in performance achieved by the two delay-aware versions of SAC are comparable to those of the delay-unaware policy learned by Hindsight Critics. In other words, SAC's sensitivity to delayed observation is comparably solved with traditional state augmentation methods and with our method. As a consequence, the combination of both, i.e. delay-aware Hindsight Critics, achieve the highest overall performances.

\section{Discussion}

\subsection{Hindsight Critics can help but not fully guide delayed actors}

The central question behind this research was wether augmenting RL critics with hindsight was enough to compensate for the performance degradation of delayed actors. To answer this question, we modified SAC, a high-performing algorithm in unperturbed conditions but particularly sensitive to delays. Our method, Hindsight Critics, showed a consistent improvement relative to SAC, retaining the zero-delay performance for moderate delays across several environments (see Fig.~\ref{fig:hc_vs_sac}). However, we found that larger delays were too severe be mitigated by delayed actors alone. In sum, we have shown that Hindsight Critics can retain SAC's zero-delay performance and increase its robustness under moderate delays, but other delay-aware augmentations are required to compensate for large delays.

\begin{table}[!t]
\caption{AUC scores for the delay-aware versions (history stacking and forward model) of Hindsight Critics (HC), PPO, and SAC, across multiple environments. The highest score for each delay-aware version and environment is highlighted in \textbf{bold}.} 
\label{tab:aware}
\centering
\begin{tabular*}{\textwidth}{l@{\extracolsep{\fill}}cccccc}
\\
\toprule
& \multicolumn{3}{c}{history stacking} & \multicolumn{3}{c}{forward model} \\
\cmidrule(r){2-4} \cmidrule(r){5-7} 
Environment     & PPO & SAC & \textbf{HC} & PPO & SAC & \textbf{HC} \\
\midrule
\texttt{Pendulum-v1} & .243 & .813 & \textbf{.859} & .130 & .740 & \textbf{.817}\\
\texttt{InvertedPendulum-v5}  & \textbf{.219} & .023 & .024 & \textbf{.228} & .023 & .024 \\
\texttt{InvertedDoublePendulum-v5}  & .036 & .046 & \textbf{.049} & .047 & .046 & \textbf{.048} \\
\texttt{Reacher-v5}  & .757 & .862 & \textbf{.875} & .714 & \textbf{.859} & .857 \\
\texttt{HalfCheetah-v5}  & .128 & .187 & \textbf{.406} & .164 & .349 & \textbf{.539} \\
\texttt{Hopper-v5} & \textbf{.121} & .034 & .039 & \textbf{.289} & .034 & .031 \\
\texttt{Walker2d-v5} & \textbf{.159} & .015 & .017 & \textbf{.149} & .013 & .038\\
\texttt{Swimmer-v5}  & .309 & .233 & \textbf{.341} & .308 & .233 & \textbf{.323} \\
\bottomrule
\end{tabular*}
\end{table}

\subsection{Real-world delays can be stochastic and unpredictable}

The main limitation of our work is that, since the actor's observations are not modified, the delayed actors must implicitly anticipate the environment's future states to learn a successful policy. This is only feasible under the assumption that delays are fixed and homogeneous. Real-world systems must deal with heterogeneous delays and jittery communication latencies that make such an anticipatory policy too brittle. However, we found that Hindsight Critics can complement delay-aware methods, in particular history stacking and forward models, that can alleviate the consequence of delay unpredictability. In future work, we will explore how Hindsight Critics compares with SAC and other RL algorithms in terms of robustness to stochastic delays.

\subsection{Hindsight Critics can generalise to other POMDPs}

The formalism introduced here can be readily generalised to other POMDPs provided that state estimates can be recovered from completed trajectories. In such cases, the actor can learn a policy conditioned on the observations $x_t$ and the critics learn value functions for the triplets of observation $x_t$, state estimate $\tilde{s}_t$, and action $a_t$. The obvious difficulty lies in obtaining state estimates from partial observations. The original asymmetric actor-critic \cite{pinto2017asymmetric} assumes privileged information for the critic by providing the true states, effectively eliminating the problem of estimating the states. Instead, we postulate that the state can be approximated using complete rollouts by means of a hindsight state estimator function $\mathcal{H}(x_{0:T}, a_{0:T}, t)$.

The hindsight function may use the observations and actions from an entire rollout but it must approximate the true state of the environment at the moment of the agent's decision making to enhance the critic's value estimation. As such, it should prevent any leakage of rewards or consequences of the action $a_t$. The choice of hindsight function should be informed by the limits imposed by the partial observability. For delayed observations, the natural choice used in this paper is $\mathcal{H}(x_{0:T}, a_{0:T},  t) = x_{t+\tau}$. For a visual agent with a limited field-of-view, $\mathcal{H}$ could encode the entire arrangement of objects in a room composed from multiple sequential observations. In a more general setup, if the true state $s_t$ is available, $\mathcal{H}$ can be a neural network with parameters $\omega$ trained to minimise a state estimation error
\begin{equation}
    \omega^* = \arg \min_\omega \;\lvert \mathcal{H}_\omega(x_{0:T}, a_{0:T}, t) - s_t \rvert \,.
\end{equation}

Such learned hindsight functions will be implemented in future work.

\section{Conclusion}

We have presented Hindsight Critics, a RL algorithm that extends SAC with an asymmetry in the information available to the actor and the critics: while the actor must act using partial observations, the critics have access to state estimates constructed via hindsight from a replay buffer. When applied to the problem of observation delays, Hindsight Critics partially alleviated the delay degradation of SAC and showed high performance when used in combination with other delay mitigation methods. Our work showed that hindsight can provide an effective mechanism for dealing with delays without requiring modifications to the agent's observations during deployment.

%
% --- Credits ---
%

\begin{credits}
\subsubsection{\ackname} This research includes computations using the computational cluster Katana \cite{katana} supported by Research Technology Services at UNSW Sydney.
\subsubsection{\discintname} The authors have no competing interests to declare that are
relevant to the content of this article.
\end{credits}

\bibliographystyle{splncs04}
\bibliography{1_REFERENCES}

\end{document}